\documentclass[10pt]{article} 
\usepackage[preprint]{rlc}

\usepackage{amssymb}            
\usepackage{mathtools}          
\usepackage{mathrsfs}           
\mathtoolsset{showonlyrefs}     
\usepackage{graphicx}           
\usepackage{subcaption}         
\usepackage[space]{grffile}     
\usepackage{url}                

\usepackage{array}
\usepackage{algorithm}
\usepackage{etoolbox}
\let\classAND\AND
\let\AND\relax
\usepackage{algorithmic}

\let\AND\classAND
\AtBeginEnvironment{algorithmic}{\let\AND\algoAND}
\usepackage{tabularx}
\usepackage{amsmath}
\usepackage{amsthm}
\usepackage{booktabs}

\title{CURO: Curriculum Learning for Relative Overgeneralization}

\author{Lin Shi\thanks{Equal contribution.} \\
    linshi1121@gmail.com \\
    University of Michigan, Ann Arbor 
    \And
    Qiyuan Liu$^*$ \\
    uceeql2@ucl.ac.uk\\
    University College London 
    \AND
    Bei Peng \\
    bei.peng@liverpool.ac.uk \\
    University of Liverpool
    }

\begin{document}

\maketitle

\begin{abstract}
Relative overgeneralization (RO) is a pathology that can arise in cooperative multi-agent tasks when the optimal joint action's utility falls below that of a sub-optimal joint action. 
RO can cause the agents to get stuck into local optima or fail to solve cooperative tasks requiring significant coordination between agents within a given timestep.
In this work, we empirically find that, in multi-agent reinforcement learning (MARL), both value-based and policy gradient MARL algorithms can suffer from RO and fail to learn effective coordination policies. 
To better overcome RO, we propose a novel approach called curriculum learning for relative overgeneralization (CURO).
To solve a target task that exhibits strong RO, in CURO, we first fine-tune the reward function of the target task to generate source tasks to train the agent.
Then, to effectively transfer the knowledge acquired in one task to the next, we use a transfer learning method that combines value function transfer with buffer transfer, which enables more efficient exploration in the target task. 
CURO is general and can be applied to both value-based and policy gradient MARL methods. 
We demonstrate that, when applied to QMIX, HAPPO, and HATRPO, CURO can successfully overcome severe RO, achieve improved performance, and outperform baseline methods in a variety of challenging cooperative multi-agent tasks. 
\end{abstract}

\section{Introduction}
\label{sec:introduction}

Cooperative multi-agent reinforcement learning (MARL) holds great promise in solving real-world multi-agent problems, such as multi-robot search and rescue~\citep{queralta2020collaborative} and traffic light control~\citep{calvo2018heterogeneous}. 
Under the paradigm of centralized training with decentralized execution (CTDE)~\citep{oliehoek2008optimal,kraemer2016multi}, VDN~\citep{sunehag2017value} and QMIX~\citep{rashid2018qmix} are two popular value-based MARL algorithms that both aim to efficiently learn a centralized but factored joint action-value function. To ensure consistency between the centralized and decentralized policies, VDN and QMIX both represent the joint action-value function as a monotonic combination of per-agent utilities. However, this monotonicity constraint prevents VDN and QMIX from representing joint action-value functions that are \textit{nonmonotonic}~\citep{mahajan2019maven}, i.e., an agent’s ordering over its own actions depends on other agents’ actions.

Recent work~\citep{gupta2021uneven} shows that both VDN and QMIX are prone to a multi-agent pathology called \textit{relative overgeneralization} (RO)~\citep{wei2016lenient},
which hinders the accurate learning of the joint action-value functions.
In cooperative tasks, RO can arise when matching an agent's available actions with arbitrary random actions from the collaborating agents results in a sub-optimal action receiving the highest utility estimate. 
This can cause the agents to get stuck into local optima or fail to solve cooperative tasks requiring significant coordination between agents.

\begin{figure}[t]
\centering
\includegraphics[width=0.45\columnwidth]{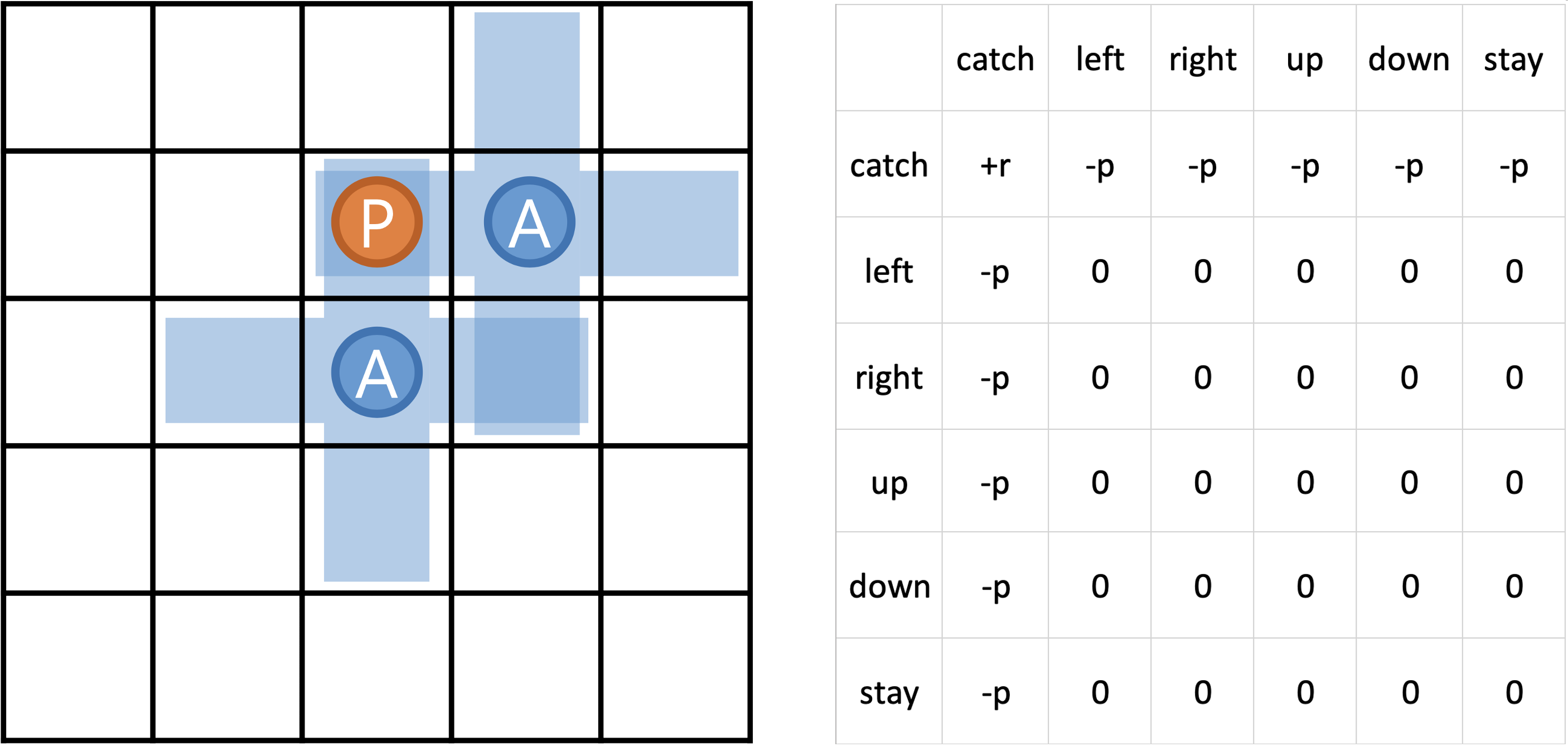}
\caption{The partially-observable predator-prey task, where two agents are rewarded when they execute the \textit{catch} action simultaneously and punished when one attempt it alone \protect\citep{son2019qtran}.}
\label{fig:predator_prey}
\end{figure}

An intuitive example illustrating the RO pathology is the partially-observable predator-prey task~\citep{son2019qtran,bohmer2020deep} as shown in Figure~\ref{fig:predator_prey}.
In this task, 2 agents (predators) need to coordinate with each other to capture a prey.
If both agents surround the prey and execute the \textit{catch} action simultaneously, a prey is caught and a positive reward of magnitude $r$ is given. However, if only one surrounding agent performs the \textit{catch} action, a negative reward of magnitude $p$ is given. Otherwise, no reward is given. 
When an agent estimates the utility of taking the catch action, it typically considers arbitrary action that the other agent might take and estimates the average utility, since it does not know the other agent's action.  
The miscoordination penalty term $p$ can then dominate the average value estimated by each agent’s utility. This can lead the agents to choose “safe” actions that are likely to tempt them away from the optimal joint action.
Agents eventually converge to sub-optimal solutions since there is not enough positive experience to help agents learn an accurate estimate of the optimal joint action's Q-value. RO can therefore be detrimental to many cooperative multi-agent tasks that require extensive coordination among agents.

Recent value-based MARL methods such as WQMIX~\citep{rashid2020weighted} and QPLEX~\citep{wang2021qplex} have demonstrated some success in solving the predator-prey task exhibiting RO as mentioned above, however, we empirically find that they can still fail to solve cooperative tasks that exhibit strong RO (e.g., the same predator-prey task but with a high miscoordination penalty $p$). 
Furthermore, although RO has mostly been investigated in value-based MARL approaches, our experimental results show that RO pathology can also arise in multi-agent policy gradient methods, preventing agents from improving their policies efficiently or escaping local optima. 
Therefore, developing reliable MARL methods for tackling RO is a particularly
important problem. 


The main idea of this paper is to combine curriculum learning~\citep{bengio2009curriculum} with MARL to better overcome RO.
We propose a novel approach called CUrriculum learning for Relative Overgeneralization (CURO), which can be applied to both value-based and policy gradient MARL algorithms.
Our key insights is that, to solve a target task exhibiting strong RO, we can start with simpler tasks that do not exhibit RO or exhibit mild RO, and then leverage these positive experiences to guide the agent to explore optimal actions more efficiently when learning harder tasks with increased severity of RO.  

In CURO, to create a curriculum (i.e., a sequence of source tasks with increasing difficulty), we fine-tune the reward function of the target task to control the probability of RO occurring, in order to find suitable source tasks that are as similar as possible to the target task, while not exhibiting a strong RO problem. 
To effectively transfer the knowledge acquired in one task to the next in the curriculum, with off-policy algorithms, we use a transfer learning method that combines \textit{value function} transfer \citep{taylor2007transfer} with \textit{buffer} transfer \citep{wang2020few}. 
In value function transfer, the parameters of a value function learned in the previous task are used to initialize the value function in the next task. 
In buffer transfer, the state-action-reward experience tuples stored in the replay buffer in the previous task are used to initialize the replay buffer in the next task. 
We find that the use of buffer transfer is critical in helping overcome RO in off-policy methods, since it prevents the value function learned from the previous task from being adjusted too much when learning the new task to produce positive experiences.
For on-policy algorithms, we simply use value function/policy transfer since it is impossible to apply buffer transfer.

To validate CURO, we apply it to both value-based and policy gradient MARL methods. 
For the former, we apply CURO to QMIX and evaluate its performance in partially-observable predator-prey tasks and challenging StarCraft
Multi-Agent Challenge (SMAC)~\citep{samvelyan2019starcraft} benchmark tasks that both exhibit strong RO and require significant coordination among agents.
For the latter, we apply CURO to HATRPO and HAPPO~\citep{kuba2021trust} and evaluate their performance on the Multi-Agent MuJoCo~\citep{peng2021facmac} benchmark. 
Our experimental results show that, when applied to QMIX, HATRPO, and HAPPO, CURO can successfully overcome severe RO, achieve improved performance, and outperform baseline methods.

Our aim in this work is \textit{not} to propose new curriculum learning algorithms for MARL. The novelty of our work lies in the novel use of curriculum learning to enable a variety of MARL algorithms, including both value-based and policy gradient methods, to overcome RO more efficiently and reliably than existing methods in both discrete and continuous cooperative tasks. Additionally, CURO is general and can be readily applied to other value-based and policy gradient MARL methods. 

\section{Background and Related Work}
\label{sec:background}

We consider a fully cooperative multi-agent task in which a team of agents interacts with the same environment to achieve some common long-term goal. It can be modeled as a \textit{decentralized partially observable Markov decision process} (Dec-POMDP)~\citep{oliehoek2016concise} consisting of a tuple
\(G=\langle\mathcal{N},\mathcal{S},\mathcal{U},P,R,\Omega,O,\gamma\rangle\). Here $\mathcal{N} \equiv \{1,\dots,n\}$ denotes the finite set of agents and \(s\in\mathcal{S}\)
describes the true state of the environment. At each time step, each agent
\(a\in\mathcal{N}\)
chooses an action 
\(u^a\in\mathcal{U}\),
forming a joint action
\(\boldsymbol{u}\in\boldsymbol{\mathcal{U}}\equiv\mathcal{U}^n\).
This causes a transition in the environment according to the state transition kernel
\(P(s'|s,\boldsymbol{u}):\mathcal{S}\times\boldsymbol{\mathcal{U}}\times\mathcal{S} \rightarrow [0,1]\).
All agents share the same reward function
\(R(s,\boldsymbol{u}):\mathcal{S}\times\boldsymbol{\mathcal{U}}\rightarrow\mathbb{R}\).
\(\gamma\in [0,1)\)
is a discount factor specifying how much immediate rewards are preferred to future rewards.
Due to
\textit{partial observability},
each agent $a$ 
receives an individual partial observation
\(o^a\in\Omega\)
drawn from the observation kernel $o^a \sim O(s,a)$.
At time
\(t\),
each agent
\(a\)
has access to its action-observation history
\(\tau_t^a\in\mathcal{T}_t\equiv(\Omega\times\mathcal{U})^t\times\Omega,\)
on which it conditions a stochastic policy
\(\pi^a(u^a_t|\tau^a_t).\)
\(\boldsymbol{\tau}_t\in\mathcal{T}_t^n\)
denotes the histories of all agents. The joint stochastic policy
\(\boldsymbol{\pi}(\boldsymbol{u}_t|s_t,\boldsymbol{\tau}_t)\equiv\prod_{a=1}^n\pi^a(u_t^a|\tau_t^a)\)
induces a joint action-value function:
\(Q^{\boldsymbol{\pi}}(s_t,\boldsymbol{\tau}_t,\boldsymbol{u}_t)=\mathbb{E}[G_t|s_t,\boldsymbol{\tau}_t,\boldsymbol{u}_t]\),
where
\(G_t=\sum_{i=0}^\infty\gamma^i r_{t+i}\)
is the discounted return.

\subsection{Relative Overgeneralization}
\label{sec:RO}
In deep MARL, there are only few existing works that focus on addressing relative overgeneralization (RO). One example is the multi-agent soft $Q$-learning method proposed by~\citeauthor{wei2018multiagent}~\citeyearpar{wei2018multiagent}, which extends soft $Q$-learning~\citep{haarnoja2017reinforcement} to multi-agent settings using the CTDE paradigm. 
Another example is NUI-DDQN~\citep{palmer2019negative}, which identifies and discards episodes yielding cumulative rewards outside the range of expanding intervals. By doing this, NUI-DDQN reduces the noise introduced by the large punishments resulting from miscoordination and improves the accuracy of the estimated utility values. 
However, multi-agent soft $Q$-learning is only evaluated in simple single state games. It models the joint action of all the agents, which grows exponentially with the number of agents. 
NUI-DDQN is an independent learning method, which cannot explicitly represent complex interactions between the agents and may not converge. 
Both methods do not scale well to challenging cooperative tasks with a large number of agents and/or actions. 

More recent deep value-based MARL methods, such as WQMIX~\citep{rashid2020weighted} and QPLEX~\citep{wang2021qplex} can overcome RO to some extent. 
WQMIX introduces a weighting scheme to place more importance on the better joint actions to learn a richer class of joint action-value functions. 
QPLEX can represent all possible classes of joint action-value functions by leveraging the dueling network architecture. 
While both WQMIX and QPLEX have demonstrated some success in overcoming RO, we empirically find that they can still fail to solve cooperative tasks that exhibit strong RO. 
We still lack reliable and scalable MARL methods for tackling RO.


\subsection{Curriculum Learning}
\label{sec:CL}
In the context of machine learning, curriculum learning (CL)~\citep{bengio2009curriculum, silva2019survey, narvekar2021curriculum} is a methodology to optimize the order in which training examples is presented to the learner, so as to increase performance or training speed on one or more final tasks. 
CL has been largely used in RL to train the agent to solve tasks that may otherwise be too difficult to learn from scratch~\citep{narvekar2021curriculum}.
For RL domains, when creating a curriculum (i.e., a sequence of source tasks), the source tasks
may differ in state/action space, reward function, or transition function from the target task. 
See~\citep{taylor2009transfer} for an in-depth overview of transfer learning in RL, and~\citep{narvekar2021curriculum} for an in-depth overview of CL in RL.

While CL has been extensively used in single-agent RL to improve learning and generalization, it has received comparatively less attention from the MARL community. 
Recent works focus on leveraging CL to scale MARL methods to
more complex multi-agent problems with a large population of agents~\citep{gupta2017cooperative, long2020evolutionary}.
In this work, our aim is not to propose better algorithms for automatically generating curricula. 
We focus on using CL to enable a variety of MARL algorithms to overcome RO in a more reliable way than existing methods.

\section{Methodology}
\label{sec:method}
In this section, we present our approach called CUrriculum learning for Relative Overgeneralization (CURO) that aims to combine curriculum learning with MARL to better overcome relative overgeneralization (RO) in cooperative multi-agent tasks. 


One main problem we need to address first is how to create a sequence of source tasks with increasing difficulty that is tailored to the current ability of the learning agent. 
Consider the partially-observable predator-prey task as mentioned above, the probability that RO occurs in this task increases if we increase the magnitude of the penalty $p$ associated with each miscoordination, since the agent is more discouraged from taking the optimal individual action (i.e., the \textit{catch} action) in the future after being punished more.
If we remove the penalty associated with each miscoordination, i.e., $p=0$, the corresponding task is likely to be free from RO.
Inspired by this, to create a curriculum, we change only the reward function of the target task to control the probability of RO occurring, without modifying any other components (e.g., the state space and action space) of the target task Dec-POMDP.
More specifically, to determine the reward function of the source tasks, we generate a sequence of candidate reward functions by reducing the magnitude of the miscoordination penalty term in the reward function of the target task, while keeping other components of the reward function the same. \(<R_1, R_2, \dots, R_n>\) is the resulting sequence of candidate reward functions, which are ordered from the smallest miscoordination penalty term to the largest one.

After generating the curriculum, the next question is how to effectively transfer knowledge acquired in one task to the next.
One naive method is to use value function transfer~\citep{taylor2007transfer}, where the parameters of a value function learned in the previous task are used to initialize the value function in the next task, such that the agent learns the next task with some initial policy that is better than random exploration. 
However, when learning the target task exhibiting strong RO, the good initial policy obtained
from the previously learned task 
can quickly converge to sub-optimal due to the large number of negative sample experiences added to the replay buffer in early stage. 

\begin{algorithm}[tb]
\caption{Curriculum Learning for RO (CURO)}
\label{alg:source_task_generation}


\textbf{Input}: Target task \(M\), a sequence of candidate reward functions $<R_1, R_2, \dots, R_n>$

\begin{algorithmic}[1] 

\STATE Train an existing MARL method on $M$ until it converges and save the learned policy $\pi_{M}$

\STATE \(M^\prime \gets M\)

\STATE $\theta_{M^\prime} \gets \theta_{random}$
\STATE $\mathcal{D}_{M^\prime} \gets \emptyset$ 

\FOR{\(R^\prime\ {\rm in}\ <R_1, R_2, \dots, R_n>\)}

\STATE \(M^\prime.R\gets R^\prime\)


\STATE Set the Q-value for $M^\prime$ with weight $\theta_{M^\prime}$ and set the replay buffer for $M^\prime$ as $\mathcal{D}_{M^\prime}$

\STATE Train the same MARL method on $M^\prime$ until it converges and save the learned action-value function as $Q(s,a|\theta_{M^\prime})$ and replay buffer as $\mathcal{D}_{M^\prime}$

\STATE Initialize the $Q$-value for $M$ with weight $\theta_{M} \gets \theta_{M^\prime}$
\STATE Initialize the replay buffer for $M$ as $\mathcal{D}_{M} \gets \mathcal{D}_{M^\prime} $






\STATE Train the same MARL method on $M$ until it converges and get policy $\pi_{test}$






\IF {\(\rm AverageTestReturn(M,\pi_{test}) \gg 
\rm AverageTestReturn(M,\pi_{M})\)}

\STATE \textbf{return} $\pi_{test}$

\ENDIF
\ENDFOR
\end{algorithmic}
\end{algorithm}

To address this, we use a transfer learning method that combines value function transfer~\citep{taylor2007transfer} with buffer transfer \citep{wang2020few}. \footnote{For MARL algorithms that do not use replay buffer, we simply use value function transfer or policy transfer.}
In buffer transfer, after training the agent on the source tasks, the state-action-reward experience tuples stored in the replay buffer are used to initialize the replay buffer for the target task. 
When training on the target task, the old sample experiences in the replay memory are gradually replaced by new sample experiences generated in the target task.
In our ablation experiments, we find that the use of buffer transfer is critical in helping overcome RO in off-policy methods, since it prevents the value function learned from the previous task from being adjusted too much when learning the new task to produce positive experiences. 
Moreover, the buffer inherited from previous source task usually contains optimal joint actions, which can overlap in some states with the target task due to the same goal of catching the prey.

Pseudocode for our proposed method, which we call CUrriculum learning for Relative Overgeneralization (CURO), is shown in Algorithm~\ref{alg:source_task_generation}.  
Given a target task $M$ we are interested in solving, we first use an existing MARL algorithm to train a policy $\pi_{M}$ on $M$, which should fail to learn a good coordination policy due to the strong RO problem present in $M$.~\footnote{We use QMIX, HATRPO, or HAPPO to train the policy $\pi_{M}$ on the target task in our experiments.}  
Then, we generate a sequence of candidate reward functions \(<R_1, R_2, \dots, R_n>\) using the simple heuristic discussed above. We start with a candidate source task $M^{\prime}$ with reward function $R_1$, which is likely to be free from RO or exhibit mild RO. 
We train the MARL method on $M^{\prime}$ until it converges and save the learned action-value function $Q(s,a|\theta_{M^{\prime}})$ and replay buffer $\mathcal{D}_{M^{\prime}}$. 
Then, we move to the target task $M$ and do value function transfer and buffer transfer, initializing the action-value function with parameters $\theta_{M^{\prime}}$ and setting the initial replay buffer for $M$ as $\mathcal{D}_{M^{\prime}}$. 
We then train the same MARL method on $M$ until it converges and check if it can overcome RO and learn an effective coordination policy. 
If the average test return of the converged policy $\pi_{test}$ is substantially higher than the average test return of $\pi_{M}$ in $M$, the target task has been solved, and we return $\pi_{test}$. 
Otherwise, we need to find a new source task that is closer to the target task without exhibiting strong RO, to allow effective knowledge transfer.
We thus move to the next source task with reward function $R_2$, transfer the saved action-value function $Q(s,a|\theta_{M^{\prime}})$ and replay buffer $\mathcal{D}_{M^{\prime}}$ from the previous source task with reward function $R_1$ and repeat the above steps until we find a good $\pi_{test}$ for $M$. 
\section{Experimental Results}
\label{sec:prep}
In this section, we present our experimental results on three multi-agent domains. 
To validate CURO, we apply it to both value-based and policy gradient MARL algorithms. 
For the former, we apply CURO to QMIX~\citep{rashid2020monotonic}, an off-policy method working only for discrete cooperative tasks, and refer to our method as CURO-QMIX. 
For the latter, we apply CURO to HAPPO~\citep{kuba2021trust}, and HATRPO~\citep{kuba2021trust}, which are both on-policy methods, and refer to our methods as CURO-HAPPO and CURO-HATRPO, respectively. 
We compare these against baseline MARL algorithms including QMIX, HAPPO, HATRPO, WQMIX~\citep{rashid2020weighted}, and QPLEX~\citep{wang2021qplex}.
For evaluation, all experiments in this section are carried out with $5$ different random seeds. 
More experimental details are included in the Appendix.

\subsection{Domain 1: Cooperative Predator-Prey}

\begin{figure*}[t]
\centering
\includegraphics[width=0.8\textwidth]{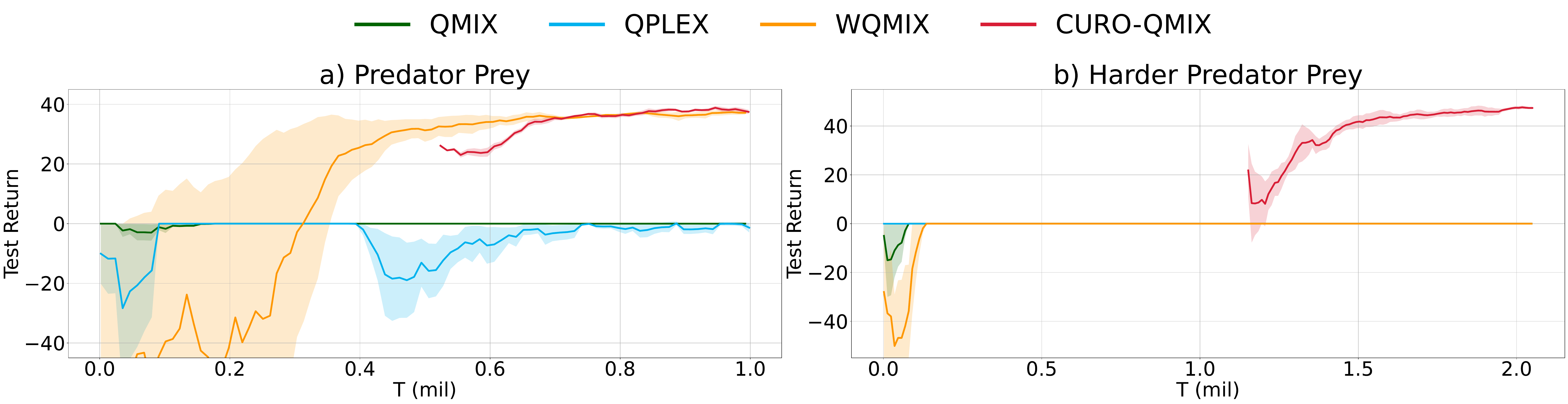}
\caption{Mean test return for QMIX, QPLEX, WQMIX, and CURO-QMIX on predator-prey with different levels of difficulty. The 95\% confidence interval is shown shaded. For CURO-QMIX, the learning curve is offset to reflect timesteps in source tasks. 
}
\label{fig:test_return_pred_prey}
\end{figure*}

We first evaluate CURO-QMIX on cooperative predator-prey tasks similar to one proposed by~\citeauthor{son2019qtran}~\citeyearpar{son2019qtran}, but significantly more complex in terms of the coordination required among agents.
We consider a partially-observable predator-prey task involving $8$ agents (predators) and $8$ prey in a $10\times10$ grid. 
A positive reward of $+10$ is given if a prey is caught and a negative reward of $-2$ is given if only one surrounding agent performs the catch action.
An agent's observation is a $5 \times 5$ sub-grid centered around it. 
The punishment for miscoordination makes this task exhibit RO. 

Figure~\ref{fig:test_return_pred_prey}(a) illustrates the median test return attained by different methods on this cooperative task during testing. We can see that QMIX fails to learn a policy that achieves any positive reward due to its monotonicity constraint, which hinders the accurate learning of the values of different joint actions. Interestingly, QPLEX also fails to solve the task despite not having any restrictions on the joint action-value functions it can represent. By contrast, both CURO-QMIX and WQMIX can successfully solve this task that exhibits RO.


We then increase the difficulty of the predator-prey task by making the capture condition stricter: each prey needs to be captured by at least three surrounding agents (rather than two surrounding agents) with a simultaneous capture action. The miscoordination penalty term remains to be $-2$. 
In addition, we increase the size of the grid world and the number of agents, with $16$ predators and $16$ prey in a $16\times16$ grid, to test the robustness of our method to RO.

As shown in Figure~\ref{fig:test_return_pred_prey}(b), CURO-QMIX can still successfully solve this task that exhibits stronger RO, whereas all other three multi-agent value-based methods fail to learn anything useful in this task. This suggests that WQMIX can struggle with cooperative tasks that exhibit strong RO, while our method can overcome RO more reliably. We emphasize that while both QPLEX and WQMIX can represent an unrestricted joint action-value function, which may allow them to overcome RO, they are not guaranteed to do so in any reasonable amount of time. This is mainly due to the simple noise-based exploration (i.e., \(\varepsilon\)-greedy action-selection strategy) used in these algorithms, which is shown to be sub-optimal in a MARL setting and can result in slow exploration and sub-optimal solutions in complex environments~\citep{mahajan2019maven}. 
By contrast, our approach can enable more efficient exploration in the target task that exhibits strong RO, by leveraging the positive sample experiences collected from previously solved simpler tasks to bias the agents towards optimal joint actions.

\subsection{Domain 2: StarCraft Multi-Agent Challenge}

\begin{figure*}[t]
\centering
\includegraphics[width=0.9\textwidth]{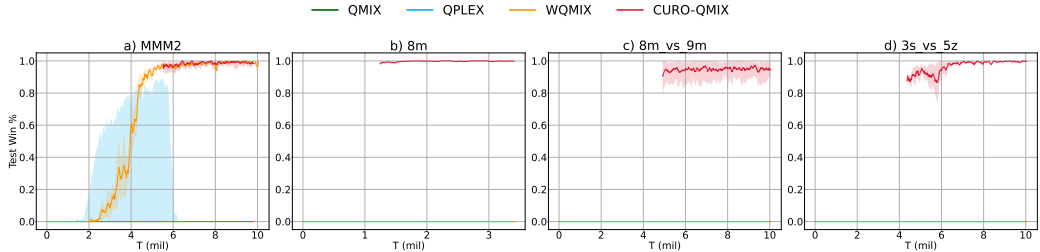}
\caption{Median test win rates for QMIX, QPLEX, WQMIX and CURO-QMIX on four different SMAC maps with negative reward scaling $p=1.0$. The $0-100\%$ percentiles are shown
shaded. For CURO-QMIX, the learning curve is offset to reflect timesteps in source tasks. }
\label{fig:test_win_rate_smac}
\end{figure*}

We also evaluate our method on StarCraft Multi-Agent Challenge (SMAC)~\citep{samvelyan2019starcraft} benchmark. SMAC consists of a set of complex StarCraft II micromanagement tasks that are carefully designed to study decentralized multi-agent control. The tasks in SMAC involve combat between two armies of units. The first army is controlled by a group of learned allied agents. The second army consists of enemy units controlled by the built-in heuristic AI. The goal of the allied agents is to defeat the enemy units in battle, to maximize the win rate. As shown by~\citeauthor{gupta2021uneven}~\citeyearpar{gupta2021uneven}, the default reward function used in SMAC does not suffer from RO as it has been designed with QMIX in mind. Specifically, each ally agent unit receives positive reward for killing/damage on enemy unites, but does not receive punishment for being killed or suffering damage from the enemy (i.e., negative reward scaling $p = 0$). Previous work~\citep{yu2021surprising} has established that a number of value-based MARL methods such as VDN and QMIX can solve almost all SMAC maps with this default reward setting.

To better evaluate our method, which aims to overcome RO that arises in many cooperative games, we change the reward function as in~\citep{gupta2021uneven}, such that ally agents are additionally penalized for losing their ally units, inducing RO in the same way as in the above predator-prey tasks. We set $p=1.0$ to equally weight the lives of ally and enemy units. Figure~\ref{fig:test_win_rate_smac} shows the test win rate attained by different MARL methods on four different SMAC maps. We can see that both QMIX and QPLEX fail to learn anything useful in all four maps tested.
WQMIX successfully learns an effective strategy that achieves a $100\%$ win rate on \textit{MMM2}. However, it fails to learn anything useful in other three maps tested. 
By contrast, CURO-QMIX can successfully learn a good policy that consistently defeats the enemy, resulting in (nearly) $100\%$ test win rate in all four scenarios. This shows that our method can overcome severe RO reliably and achieve efficient learning in challenging coordination problems. 


\subsection{Domain 3: Multi-Agent MuJoCo}

\begin{figure*}[t]
\centering
\includegraphics[width=0.9\textwidth]{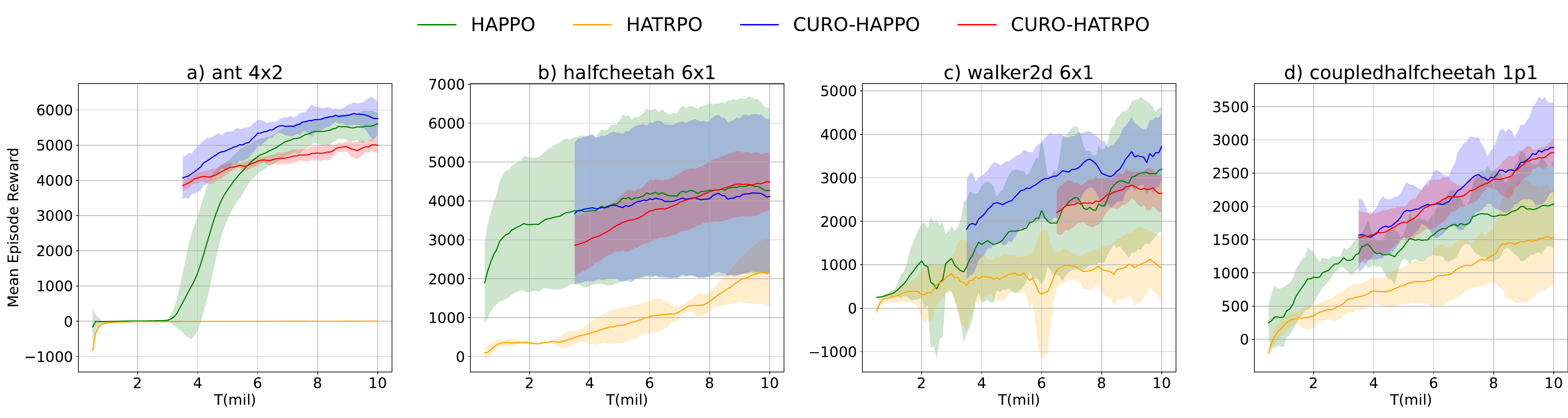}
\caption{Mean episode reward for HAPPO, HATRPO, CURO-HAPPO and CURO-HATRPO on four MaMuJoCo environments with negative reward scaling $p$. The 95\% confidence interval is shown shaded. For CURO-HAPPO and CURO-HATRPO, the learning curves are offset to reflect timesteps in source tasks. 
}
\label{fig:mamujoco}
\end{figure*}

\begin{figure*}[t]
\centering
\includegraphics[width=0.7\textwidth]{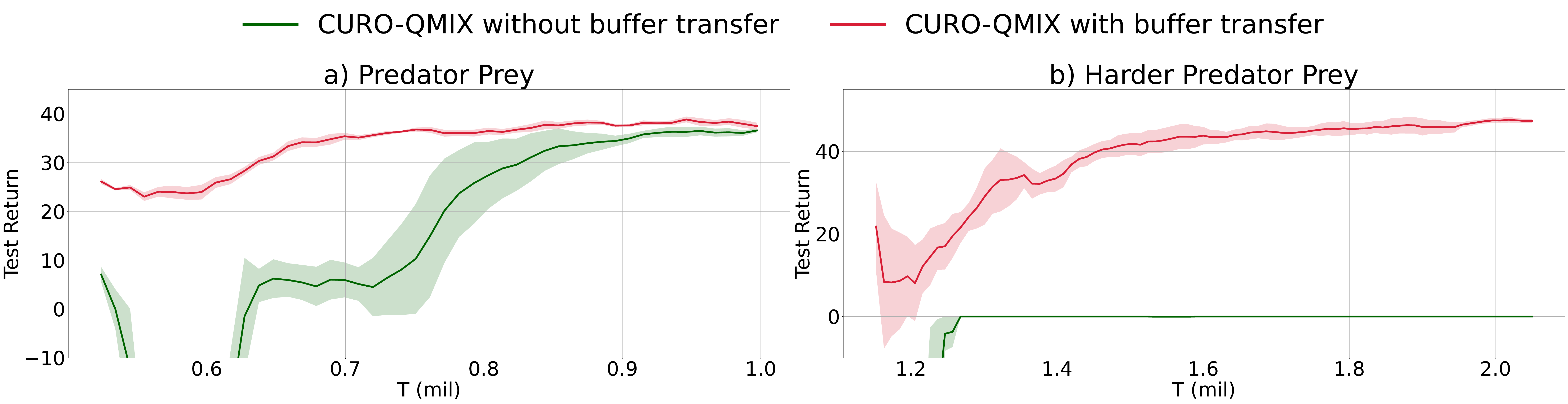}
\caption{Mean test return for CURO-QMIX with and without buffer transfer on two predator-prey tasks with different levels of difficulty. The 95\% confidence interval is shown shaded. We show only the timesteps in the target task.}
\label{fig:ablationpp}
\end{figure*}

We finally evaluate CURO on Multi-Agent MuJoCo (MaMuJoCo)~\citep{peng2021facmac}, a benchmark for continuous cooperative multi-agent robotics control. 
MaMuJoCo consists of a variety of scenarios in which multiple agents within a single robot must solve a task cooperatively. 
To better evaluate CURO's ability to overcome severe RO, we introduce modifications to the reward function similar to those in SMAC.
In MaMuJoCo tasks, agents typically receive a reward related to their displacement compared to the previous moment, which becomes negative when moving backward. 
To increase the severity of RO, we simply scale the penalty for moving backward by a factor (negative reward scaling $p$). This adjustment is designed to increase the penalty for miscoordination among agents without changing other components of rewards, such as control costs, which penalize agents for excessively large actions. 
In practice, this design is relevant in scenarios where backsliding is of high risk.

To evaluate CURO on MaMuJoCo, we apply it to HAPPO and HATRPO~\citep{kuba2021trust}, which are both on-policy multi-agent policy gradient methods, and refer to our methods as CURO-HAPPO and CURO-HATRPO, respectively.~\footnote{When training CURO-HAPPO and CURO-HATRPO on MaMuJoCo, training on each source task is stopped when it is close to convergence to improve sample efficiency.}
As shown in Figure~\ref{fig:mamujoco}, CURO-HATRPO outperforms HATRPO by a large margin in all four continuous cooperative tasks, both in terms of absolute performance and learning speed. Compared to HAPPO, HATRPO is more susceptible to severe RO problems, most notably in the Ant-4x2 environment, where CURO can give a substantial boost in training. In addition, although HAPPO does not suffer from severe RO as much as HATRPO in these environments, CURO can still bring performance improvements in most tasks, achieving higher sample efficiency and lower variance. When CURO does not bring performance improvement in HalfCheetah-6x1, it still does nor hurt the performance. 
Our results demonstrate that, other than value-based MARL methods, multi-agent policy gradient approaches can also suffer from RO, and CURO can be effectively applied to them to overcome severe RO. 

\subsection{Ablation Study}

When applying CURO to off-policy value-based MARL algorithms, we use both value function transfer and buffer transfer to transfer the knowledge between different tasks in the curriculum. 
We hypothesize that, in CURO, buffer transfer plays a crucial role in effectively reusing knowledge learned to help overcome RO. 
To investigate the effect of buffer transfer, we conduct some ablation experiments on predator-prey and SMAC. As shown in Figure~\ref{fig:ablationpp}, CURO-QMIX performs better than CURO-QMIX without buffer transfer in both predator-prey tasks.~\footnote{The results on SMAC are included in Figure 6 in the Appendix.}
This demonstrates the important role of buffer transfer in CURO when applying it to off-policy MARL methods.

\section{Conclusion}
In this work, we proposed CURO that leverages curriculum learning to better overcome RO in cooperative tasks. 
We empirically found that both value-based and policy gradient MARL approaches can suffer from severe RO and fail to learn effective coordination policies. 
Our experimental results demonstrated that,  when applied to QMIX, HAPPO, and HATRPO, CURO can successfully overcome severe RO, achieve improved performance, and outperform baseline methods in a variety of cooperative multi-agent
tasks. 
CURO is general and can be applied to other deep value-based or policy gradient MARL methods. 
One main limitation of our method is its reliance on an explicit miscoordination penalty in the reward function when fine-tuning the reward function to create a curriculum. 
Interesting directions for future work include the development of more general curriculum generation strategies 
and the application of CURO to other deep MARL methods.
\bibliography{main}
\bibliographystyle{rlc}

\appendix
\section{Related Background}
\subsection{CTDE}
\label{sec:CTDE}
We adopt the \textit{centralized training with decentralized execution} (CTDE) paradigm~\citep{oliehoek2008optimal,kraemer2016multi}.
During training, our learning algorithm has access to the true state $s$ of the environment and every agent's action-observation history $\tau^a$. 
However, during execution, each agent's decentralized policy $\pi^a$ can only condition on its own local action-observation history $\tau^a$.
This approach can use extra information that is only available during training and also enable agents to freely share their observations and internal states during training, which can greatly improve the efficiency of learning~\citep{foerster2016learning,foerster2018counterfactual}.

\subsection{Value Function Factorization}
\label{sec:valuefunctionfactorization}
Value function factorization~\citep{koller_computing_1999} has been widely employed in value-based MARL algorithms.
VDN~\citep{sunehag2017value} and QMIX~\citep{rashid2018qmix} are two representative examples that both learn a centralized but factored action-value function $Q_{tot}$, using CTDE. 
They are both $Q$-learning algorithms for cooperative MARL tasks with discrete actions. 
To ensure consistency between the centralized and decentralized policies, VDN and QMIX factor the joint action-value function $Q_{tot}$ assuming additivity and monotonicity, respectively. 
More specifically, VDN factors $Q_{tot}$ into a sum of the per-agent utilities: $Q_{tot}(\boldsymbol{\tau},\boldsymbol{u};\boldsymbol{\theta})=\sum_{a=1}^n Q^a(\tau^a,u^a;\theta^a)$.
QMIX, however, represents $Q_{tot}$ as a continuous monotonic mixing function of each agent's utilities: $Q_{tot}(\boldsymbol{\tau},\boldsymbol{u},s;\boldsymbol{\theta}, \phi)=f_{\phi}\bigl(s,Q^1(\tau^1,u^1;\theta^1),\dots,Q^n(\tau^n,u^n;\theta^n)\bigr)$, where
$\frac{\partial f_{\phi}}{\partial Q^a} \geq 0, \forall a
\in \mathcal{N}$.
Here $f_{\phi}$ is approximated by a monotonic mixing network, parameterized by $\phi$.
The weights of the mixing network are constrained to be non-negative to guarantee monotonicity. 
These weights are produced by separate \textit{hypernetworks}, which condition on the full state $s$. This can allow $Q_{tot}$ to depend on the extra state information in nonmonotonic ways. 
The joint action-value function 
\(Q_{tot}\)
can be trained using Deep Q-Networks (DQN) ~\citep{mnih2015human}. The greedy joint action in VDN and QMIX can be computed in a decentralized fashion by individually maximizing each agent’s utility $Q^a$.

\section{Experimental Setup}

All value-based MARL algorithms are implemented in the PyMARL2 framework~\citep{hu2021rethinking}, where the baseline QMIX is fine-tuned for better performance.
For SMAC, 
we use the latest version SC2.4.10. Note that performance is not always comparable across versions.

For both SMAC and predator-prey, the sequence of negative reward scaling $p$ in source tasks selected by CURO is $<0>$.
All experiments use \(\varepsilon\)-greedy action exploration strategy. In CURO-QMIX, \(\varepsilon\) does not reset for each transfer learning step. Instead, it follows its predetermined annealing schedule, linearly decaying from $1.0$ to $0.05$ over $400k$ time steps. The replay buffer for SMAC and predator-prey contains the most recent $5000$ and $1000$ episodes respectively. For all experiments, 8 rollouts for parallel sampling are used to obtain samples from the environments.
We sample batches of $128$ episodes uniformly from the replay buffer and train on fully unrolled episodes.
All the agent networks are the same as those in QMIX~\citep{rashid2020monotonic}.
For QMIX, QPLEX, and WQMIX, all the hyperparameters are the same as the default values in PyMARL2~\citep{hu2021rethinking}.
All target networks are periodically updated every $200$ training steps. 
All networks are trained using Adam optimizer with learning rate $0.001$.
We set the discount factor \(\gamma\) = 0.99 for all experiments. 

The experiments of MaMuJoCo are run in the HARL framework \citep{zhong2023heterogeneousagent,liu2023maximum}. The negative reward scaling factor and the curriculum designed for each environment are shown in Table ~\ref{curriculum}. For CoupledHalfCheetah environment, negative reward scaling is applied only if the average displacement of two agents is negative.
Training on each source task is early stopped when it is close to convergence to improve sample efficiency. In Ant, HalfCheetah, CoupledHalfCheetah, each training on source tasks is set to 3 million time steps. In Walker2d, two source tasks are set up for HATRPO. It is successively trained 3 million time steps on each of them.

The hyperparameters used are shown in Table ~\ref{commonhyperparamamujoco},~\ref{hyperparamamujocohappo},~\ref{hyperparamamujocohatrpo}. They are set according to the fine-tuned hyperparameters in HARL, only with minor differences.

\section{Ablation Study}

\begin{figure*}[t]
\centering
\includegraphics[width=1.0\textwidth]{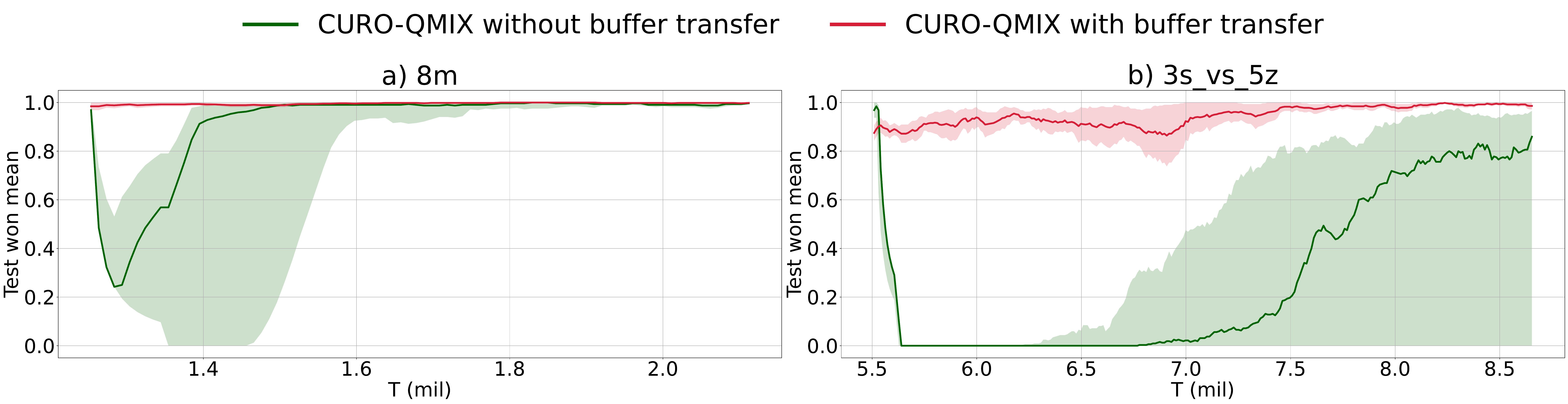}
\caption{Median test return for CURO-QMIX with and without buffer transfer on two SMAC maps \textit{8m} and \textit{3s vs 5z} with negative reward scaling $p=1$. The $0-100\%$ percentiles is shown shaded. We show only the timesteps in the target task.}
\label{fig:ablationsc}
\end{figure*}

To investigate the effect of buffer transfer, we conduct some ablation experiments on the SMAC domain.
As shown in Figure~\ref{fig:ablationsc}, CURO-QMIX appears better than CURO-QMIX without buffer transfer on both SMAC maps tested. 
CURO-QMIX without buffer transfer exhibits larger variance and instability compared to CURO-QMIX.  
This demonstrates the important role of buffer transfer in CURO when applying it to off-policy value-based MARL methods to overcome RO.

\section{Additional Experiments}

\begin{figure*}[t]
\centering
\includegraphics[width=1.0\textwidth]{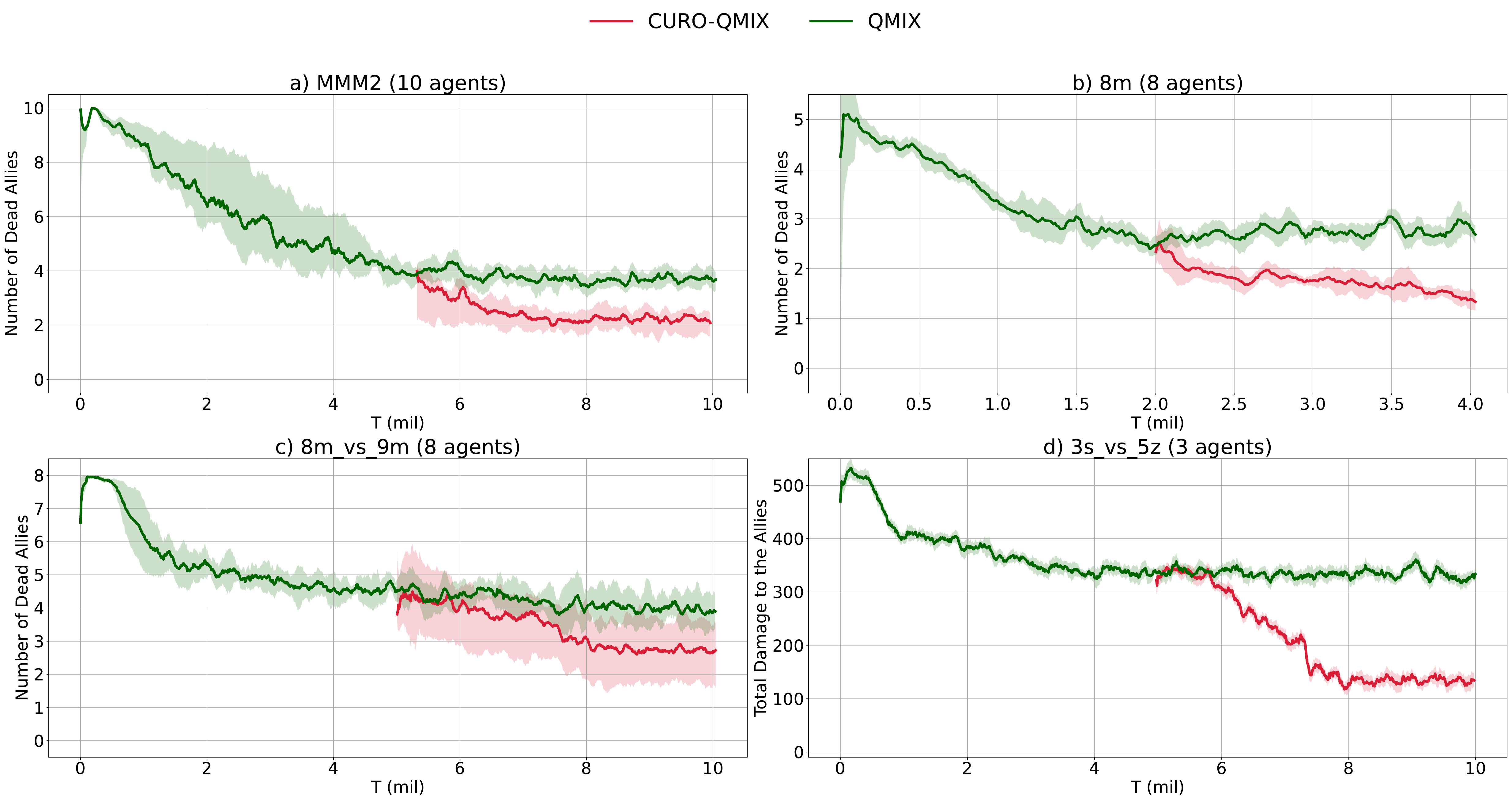}
\caption{Number of dead allies ((a)-(c)) or total damage to the allies (d) for CURO-QMIX (on SMAC maps with $p=1.0$) and QMIX (on SMAC maps with $p=0$ during testing). The $0-100\%$ percentiles is shown shaded. For CURO-QMIX, the learning curve is offset to reflect timesteps in source tasks.}
\label{fig:dead_allies_smac}
\end{figure*}

We also examine the casualties condition of ally agents resulted from the policies learned by CURO-QMIX (on SMAC maps with $p=1.0$) and QMIX (on SMAC maps with $p=0$), respectively. Figures~\ref{fig:dead_allies_smac}(a)-(c) illustrate the number of dead allies in each episode during testing on maps \textit{MMM2}, \textit{8m}, and \textit{8m\_vs\_9m}. 
For CURO-QMIX, we evaluate the number of dead allies on SMAC maps with $p=1.0$. For QMIX, we use the original SMAC setting where $p=0$ since QMIX fails to learn anything useful in SMAC maps with $p=1.0$.
For map \textit{3s\_vs\_5z}, we evaluate the total damage value to the allies in each episode as in Figure~\ref{fig:dead_allies_smac}(d), due to the low number of dead agents in this map. 
We see that, compared to QMIX, CURO-QMIX learns a more conservative policy with less total damage taken by the ally agents, while achieving nearly $100\%$ test win rate. 
This demonstrates that solving difficult tasks with RO can lead to an even better policy with distinct characteristics.

\begin{table}[tb]
    \begin{center}
        \begin{tabular}{lrrr}
            \multicolumn{1}{l}{\bf Scenario} & \multicolumn{1}{l}{\bf Target task $p$} & \multicolumn{1}{l}{\bf HAPPO $p$}  & \multicolumn{1}{l}{\bf HATRPO $p$} \\
            \hline \\
            Ant  & 30     &$<1>$    &$<1>$   \\
            HalfCheetah         & 30       &$<1>$    &$<1>$   \\
            Walker2d            & 100      &$<1>$    &$<1, 30>$   \\
            CoupledHalfCheetah  & 30       &$<1>$    &$<1>$   \\
        \end{tabular}
    \end{center}
    \caption{The negative rewarding scaling factor $p$ set in target task and the $p$ selected by CURO in each source task for HAPPO and HATPRO in MaMuJoCo}
    \label{curriculum}
\end{table}

\begin{table}[tb]
    \begin{center}
        \begin{tabular}{lr|lr|lr}
            \bfseries Hyperparameter & \bfseries Value & \bfseries Hyperparameter & \bfseries Value & \bfseries Hyperparameter & \bfseries Value \\
            \hline \\
            num mini-batch & 1 & max grad norm & 10 & std x coef & 0.5\\
            ppo epoch & 5 & gamma & 0.99 & std y coef & 1 \\
            critic epoch & 5 & gae lambda & 0.95 & training threads & 4 \\
            value loss coef & 1 & huber delta & 10 &rollout threads & 20\\
            action aggregation & prod & activation & ReLU & actor network & mlp \\
            gain & 0.01 & critic lr & 5e-3 & ls step & 10 \\
            actor lr & 5e-3 & optimizer & Adam & accept ratio & 0.5 \\
            optim eps & 1e-5 & weight decay & 0 & backtrack coef & 0.8 \\
        \end{tabular}
    \end{center}
    \caption{Common hyperparameters used for HAPPO and HATRPO in MaMuJoCo experiments}
    \label{commonhyperparamamujoco}
\end{table}


\begin{table}[tb]
    \begin{center}
        \begin{tabular}{lrrr}
            \multicolumn{1}{l}{\bf Scenario} & \multicolumn{1}{l}{\bf clip param}   & \multicolumn{1}{l}{\bf entropy coef}   &\multicolumn{1}{l}{\bf mlp hidden size}\\
            \hline \\
            Ant & 0.1 & 0 & [128, 128] \\
            HalfCheetah & 0.2 & 0.01 & [128, 128] \\
            Walker2d & 0.1 & 0.01 & [128, 128] \\
            CoupledHalfCheetah & 0.2 & 0.01 & [64] \\
        \end{tabular}
    \end{center}
    \caption{Different hyperparameters used for HAPPO in MaMuJoCo experiments}
    \label{hyperparamamujocohappo}
\end{table}

\begin{table}[tb]
    \begin{center}
        \begin{tabular}{lr}
            \multicolumn{1}{l}{\bf Scenario} & \multicolumn{1}{l}{\bf kl threshold} \\
            \hline \\
            Ant & 0.01 \\
            HalfCheetah & 0.01  \\
            Walker2d & 0.005  \\
            CoupledHalfCheetah & 0.005 \\
        \end{tabular}
    \end{center}
    \caption{Different hyperparameters used for HATRPO in MaMuJoCo experiments}
    \label{hyperparamamujocohatrpo}
\end{table}

\end{document}